\pdfoutput=1

\documentclass[twoside,11pt,a4paper]{article}

\usepackage[
backend=bibtex,
style=numeric
]{biblatex}
\usepackage{color}
\definecolor{red}{RGB}{255,0,0}
\usepackage[hmarginratio=1:1,top=25mm,columnsep=20pt]{geometry} 
\usepackage{float} 
\usepackage{hyperref} 

\usepackage{paralist} 

\usepackage{authblk}

\date{\vspace{-5ex}}

\title{\vspace{-15mm}\fontsize{18pt}{10pt}\selectfont\textbf{Convolutional Architecture Exploration for Action Recognition and Image Classification}} 

\author[1,2]{JT Turner\thanks{Student Contractor at Naval Research Laboratory from \textbf{Co}gnitive \textbf{R}obotics \textbf{a}nd \textbf{L}earning lab at University of Maryland, Baltimore County}}
\author[1]{David Aha}
\author[1]{Leslie Smith}
\author[2]{Kalyan Moy Gupta}
\affil[2]{Knexus Research Corporation; 

174 Waterfront Street Suite 310; National Harbor, MD 20745}
\affil[1]{Navy Center for Applied Research in Artificial Intelligence;

Naval Research Laboratory (Code 5514); Washington, DC 20375}

\begin{document}
\maketitle 
\begin{abstract}

\textbf{C}onvolutional \textbf{A}rchitecture for \textbf{F}ast 
\textbf{F}eature \textbf{E}ncoding (CAFFE) \cite{jia2014caffe} 
is a software package for the training, classifying, and feature extraction of images. 
The UCF Sports Action dataset is a widely used machine learning dataset that has 200 
videos taken in 720x480 resolution of 9 different sporting activities: diving, golf, swinging, kicking, lifting, horseback riding, running, skateboarding, swinging (various  gymnastics), and walking.
In this report we report on a caffe feature extraction pipeline of images taken from the videos of
the UCF Sports Action dataset. 
A similar test was performed on overfeat, and results were inferior to caffe. 
This study is intended to  explore the architecture and hyper parameters needed for effective \textbf{static} analysis of action in videos and classification over a variety of image datasets.

\end{abstract}
\section{Introduction}

Traditional action recognition focuses on temporal differences between frames in a video\cite{Cheng2015action}, 
and the movement of features extracted with various algorithms such as SIFT\cite{lowe2004distinctive} or HOG\cite{dalal2005histograms}. 
By tracking the motion of human and objects in the movie
between frames, the fully connected layers of the network can 
make predictions about what actions are being performed (i.e. a hand moving 
quickly into a white ball in a sports dataset would probably be “spiking the volleyball”). 
In an attempt to gain further knowledge of the UCF sports database, as well as caffe, 
we classified images not based upon temporal differences but instead by viewing single frame images. 
The strengths and weaknesses to this  approach are shown in the results section. 

A large factor in the success or failure of any deep neural  network is the proper
architecture set with the proper hyper parameters values \cite{chatfield2011devil}\cite{chatfield2014return}. 
This is especially true with convolutional neural networks which depend upon the architecture 
to detect edges and objects in the same way the human visual cortex does. 
Using well known image datasets such as VOC, Caltech, Stanford dogs, etc. we tested classification of images in a more traditional way and explored the tricks of the trade in
architecture and hyper parameters.

\subsection{UCF Sports Action Dataset}

The UCF Sports dataset\cite{ni2008epitomic} contains 200 videos 
of different sporting activities, and provides both the .mpeg video, and .jpg images taken at uniform  frames in the video.
We divided the video image data into training and testing splits by
randomization of video numbers, to make sure that all of the image data from a
specific video was in the same set (so we would not train on half a video and test on the
other half). 

\begin{table}[H]
\caption{UCF Sports Action Dataset}
\centering
\begin{tabular}{|l|l|l|}
\hline
Sport/Activity & Training Videos & \# Testing Videos \\
\hline
Diving & 13 & 3 \\
Golf Swinging & 21 & 4 \\
Horseback Riding & 11 & 3 \\
Kicking & 21 & 4 \\
Lifting & 12 & 3 \\
Running & 12 & 3 \\
Skateboarding & 12 & 3 \\
Swinging (Gymnastics) & 28 & 7 \\
Walking & 17 & 5 \\
\hline
\end{tabular}
\label{table:splits}
\end{table}

\subsection{Additional Datasets}
\label{subsection:datasets}
The later part of this paper focuses more on single image classification, 
which identifies objects in the image (such as a soccer ball or dumbbell) instead of recognition of an action from the video(such as kicking or weightlifting). 
The datasets used are:
\begin{compactitem}
\item\textbf{Caltech101}\cite{fei2007learning}- Created in 
2003, containing 101 different classes of objects composing a
9,146 image corpus. Examples of classes: accordion, chandelier, hedgehog.
\item\textbf{Caltech200}\cite{WelinderEtal2010}- Created in 
2011, containing 200 different classes of bird species composing a 11,788 image corpus.
Examples of classes: crested auklet, cardinal, yellow warbler.
\item\textbf{Caltech256}\cite{griffin2007caltech}- Created in 
2007, containing 256 different classes of everyday objects composing a 30,607 image corpus.
Similar to Caltech101 but more difficult. 
Example objects include: AK47, Calculator, Sushi.
\item\textbf{Olivetti Faces}- Created in 1992, containing 40 
different faces composing a 400 image corpus. 
Faces in images reflect a variety of persons: Male and females, young and old, with and without glasses. 
Racial identification hard to tell from gray scale images. 
This dataset was provided by the AT\& T Laboratories at Cambridge. 
\item\textbf{Stanford dogs}
\cite{KhoslaYaoJayadevaprakashFeiFei_FGVC2011}- Created in 
2011, containing 120 different breeds of dogs composing a 20,580 image corpus. 
Example dogs include: basset hound, rottweiler, Siberian husky.
\item\textbf{PASCAL VOC2012}\cite{pascal-voc-2012}- Created in 
2012, containing 20 different classes composing a 11,540 image corpus. 
Not the entire image corpus is used since some of the images contain more than one class.
Example classes include: airplane, bird, car.
\end{compactitem}


\section{Recognition Architecture}

Our action recognition pipeline has two algorithms that are used for classification:
convolutional neural networks, and support vector machines. 
The convolutional neural network is an 8 layer network that was 
trained on the Imagenet dataset\cite{ILSVRCarxiv14} until
performance surpassed Krizhevsky's ConvNet\cite{NIPS2012_4824}.

The more typical architecture for classification with an convolutional neural network such as Caffe is to include an "accuracy" layer.
Input to the accuracy layer is the output from a fully connected layer and the test or validation data labels (accuracy layers can follow any or all fully connected layers).  
Caffe runs the test data through the networks at specified iterations and prints the accuracy based on the number of times the network predicted the correct labels for the test data.
Future work includes comparing the two algorithm architecture used in this study with the accuracies computed directly by the network.

\subsection{Convolutional Neural Network Architecture}
\label{subsection:architecture}
The layers of the stock imagenet CNN model are as follows:

\begin{table}[H]
\caption{Pre-trained Imagenet 23 layer convolutional architecture}
\centering
\begin{tabular}{|l|l|l|l|l|}
\hline
Layer & Type & Number Kernels & \# Kernel Size & Stride \\
\hline
1 & Convolutional & 96 & $11\times11$ & $4\times4$\\
2 & Rectified Linear Unit & N/A & N/A & N/A\\
3 & Max Pooling & N/A & $3\times3$ & $2\times2$\\
4 & Local Response Normalization & N/A & $5\times5$ & 
$1\times1$\\
5 & Convolutional & 256 & $5\times5$ & $1\times1$\\
6 & Rectified Linear Unit & N/A & N/A & N/A\\
7 & Max Pooling & N/A & $3\times3$ & $2\times2$\\
8 & Local Response Normalization & N/A & $5\times5$ & 
$1\times1$\\ 
9 & Convolutional & 384 & $3\times3$ & $1\times1$\\
10 & Rectified Linear Unit & N/A & N/A & N/A\\
11 & Convolutional & 384 & $5\times5$ & $1\times1$\\
12 & Rectified Linear Unit & N/A & N/A & N/A\\
13 & Convolutional & 256 & $3\times3$ & $1\times1$\\
14 & Rectified Linear Unit & N/A & N/A & N/A\\
15 & Max Pooling & N/A & $3\times3$ & $2\times2$\\
16 & Fully Connected & 4096 & N/A & N/A\\
17 & Rectified Linear Unit & N/A & N/A & N/A\\
18 & Dropout (.5) & N/A & N/A & N/A\\
19 & Fully Connected & 4096 & N/A & N/A\\
20 & Rectified Linear Unit & N/A & N/A & N/A\\
21 & Dropout (.5) & N/A & N/A & N/A\\
22 & Fully Connected & 1000 & N/A & N/A\\
23 & Softmax & N/A & N/A & N/A\\
\hline
\end{tabular}
\label{table:architecure}
\end{table}

A brief review of the types of layers from Table~\ref{table:architecure}:
\begin{compactitem}
\item \textbf{Convolutional-} Standard n dimensional image convolution using a randomly initialzed kernel that is trained through stocastic gradient descent back propagation.
\item \textbf{Rectified Linear Unit-} Non linearity applied to the output
of convolution, defined as:
 \begin{displaymath}
   f(x) = \left\{
     \begin{array}{lr}
       0 & : x \leq 0\\
       x & : x > 0 
     \end{array}
   \right.
\end{displaymath} 
\item \textbf{Max Pooling-} This takes all of
the signals from the kernel (all are $3\times3$ in our case) and outputs
only the maximum value
\item \textbf{Local Response Normalization-} Aids convolutional neural
networks in learning \cite{NIPS2012_4824} by normalizing over local regions. 
Given the input activity of a neuron as $a^i_{x,y}$ we define the response normalized output as follows:
\[
b^i_{x,y} = \frac{a^i_{x,y}}{\bigg(k + \alpha\sum_{j=max(0, i-\frac{n}
{2})}^{min(N-1, i+\frac{n}{2})}(a^j_{x, y})^2\bigg)^\beta}
\]
where $k, n, \alpha, \beta$ are hyper parameters determined by using the validation set.
\item \textbf{Fully Connected-} Standard fully connected perceptron type
feed forward layer.
\item \textbf{Dropout-} A 50\% dropout rate is employed to discourage the
coadaption of feature detectors, as suggested in \cite{DBLP:journals/corr/abs-1207-0580}. 
This randomly zeroes out half of the values.
\item \textbf{Softmax-} Softmax function used for standard logistic 
regression and gradient descent in assigning class label.
\end{compactitem}

To extract features and not just probabilities, we obtained the feature vector 
from the fully connected layers at levels 16 and 19 in Table~\ref{table:architecure}, which output a feature vector of size 8192 floating point numbers.

\subsection{Support Vector Machine}

A basic SVM from the weka toolkit\cite{hall2009weka} was used for classification. 
SVM Parameters are those of John Platt's implementation of Polynomial Kernel SVM \cite{hall2009weka}, exponent = 2.


\section{UCF Sports Action Dataset}

This section presents results from using the Imagenet's deep neural network architecture on images from the UCF Sports Action dataset.
All the results here are from processing individual frames from the videos, not blocks of frames.

\subsection{CAFFE with stock imagenet model}

The following is the output provided by running weka. 
Note that there are a large scalar multiple more examples in this table because we are 
classifying individual frames and not videos; a single video can contain over 200 frames (videos were taken at 10 frames per second).
Table \ref{table:confusion} contains a confusion matrix, where the columns are the image classification and rows are the true labels.

\begin{table}[H]
\caption{Confusion Matrix}
\centering
\begin{tabular}{|r||r|r|r|r|r|r|r|r|r|}
\hline
  & LFT & DIV & SKT & KCK & GYM & GLF & WLK & RUN & HRB \\
\hline
LFT &\textbf{127}&0&0&0&0&0&0&0&0\\
\hline
DIV &0&\textbf{165}&0&0&0&0&0&0&0\\ 
\hline
SKT &0&0&\textbf{161}&0&0&0&\textcolor{red}{35}&\textcolor{red}
{14}&0\\ 
\hline
KCK &0&0&0&\textbf{68}&0&0&\textcolor{red}{3}&\textcolor{red}
{20}&0\\ 
\hline
GYM &0&\textcolor{red}{63}&0&0&\textbf{485}&0&0&0&0\\ 
\hline
GLF &0&0&0&\textcolor{red}{60}&0&\textbf{120}&\textcolor{red}
{60}&0&0\\ 
\hline
WLK &0&0&0&\textcolor{red}{184}&0&0&\textbf{440}&\textcolor{red}
{3}&0\\ 
\hline
RUN &0&0&0&\textcolor{red}{132}&0&0&\textcolor{red}
{63}&\textbf{0}&0\\ 
\hline
HRB &0&0&0&\textcolor{red}{1}&0&0&\textcolor{red}
{30}&0&\textbf{149}\\ 
\hline
\end{tabular}
\label{table:confusion}
\end{table}

\subsection{Preliminary Analysis}
\label{sec:prelim}
A  Google search for performance on the UCF sports action dataset shows that
in 2013 the best performance was between 90-96\% accuracy in 
recognition\cite{harandi2013kernel}, depending on what sort of testing 
split was used. The work given in \cite{harandi2013kernel} used high 
dimensional kernels for SVM to classify actions based on motion, which
relies on video knowledge as opposed to imagery, which was the work of 
this study.
The out of the box Caffe performance was 71\%, however a combination of some or all of the following techniques should improve the performance of the CNN.
However, none of the results in this report indicate any significant improvement.
\begin{compactitem}
\item Train the convolutional neural network specifically on the same dataset to be tested. 
Right now the convolutional model was trained on the Imagenet dataset which is much
larger and contains many things that would not be seen while trying to recognize sports actions.
\item Use a single classification scheme to classify the video's action.
If a video shows 28 frames of kicking, and 20 of running (as one of the movies did),
than we would classify this video as kicking. 
This would make accuracy on that video 1/1 (100\%) as opposed to 28/48 (58.3\%).
\item Leverage temporal features. 
A time series of how the features detected in the image evolve as the video progresses would contain useful information. 
This would better capture the action being performed in the video. 
One of the  clips of running was classified as golf in all frames. 
Watching the video we realized that it was taken after a golfer had just made a difficult put, and he was running around the green fist pumping with his golf club. 
The golf club led the static image analyzer to think that it was a golf action, while if 
we saw the action of rapid leg movement than we would know that he was running.
\item Change the Network Architecture. 
The architecture that we use was trained on the gargantuan Imagenet; models take up around 275 MB of hard disk. 
This is a necessary size for performance on the Imagenet dataset with over 1,000 classes; however for the smaller UCF Sports action dataset this network maybe too large. 
While a larger network would not hurt the performance in theory, it may hurt performance by limiting how many epochs can be tested on. 
The network should be large enough to express everything it needs to in the dataset but no larger.
\end{compactitem}
This preliminary analysis was performed to gain a greater understanding of feature
extractions with caffe and of the UCF sports action database.

\subsection{Overfeat with stock Imagenet model}
Overfeat\cite{sermanet2013overfeat} was the first convolutional neural network 
that we tested because of the simplicity of setting up Overfeat to run on the CPU. 
The source code for the CPU version of Overfeat is publicly available, as well 
as binaries for the CPU and NVIDIA-GPU versions of Overfeat. 
Unfortunately the GPU binary was unstable and running this executable
caused segmentation faults after around 3 images. 

We recommend  the usage of caffe not Overfeat for the following reasons:
\begin{compactitem}
\item The GPU version of caffe is working while GPU version of Overfeat is not. 
The GPU version of caffe is an order of magnitude faster than the CPU version of Overfeat.
\item Accuracy of caffe on the above test was 71.96, while the Overfeat accuracy on the same test was 53.01.
\item Caffe models can be trained for your own datasets, while 
Overfeat alone cannot be trained without first learning the companion implementations; lua and torch.
\end{compactitem}

\subsection{CAFFE with UCF Sports trained model}
The first suggestion of the list in Section \ref{sec:prelim} is to train the network on the same dataset used in testing.
This seems intuitive; that training the network to learn on everything 
but only testing on a small subset is less preferable than to train the network on the type of data used in the testing. 

The network used in the experiments reported here is the same architecture used in section~\ref{subsection:architecture}, with the exception that layer 22 has its
size modified for the correct number of outputs. 
The architecture of the neural network is illustrated in Table~\ref{table:architecure} while layer 22 is as follows:

\begin{table}[H]
\centering
\begin{tabular}{|l|l|l|l|l|}
\hline
Layer & Type & Number Kernels & \# Kernel Size & Stride \\
\hline
22 & Fully Connected & 9 & N/A & N/A\\
\hline
\end{tabular}
\end{table}

We expect that if we were to train on this small subset and test on the same subset performance should increase. However, the results shown in Table \ref{table:training_difference} shows a significant reduction in performance.  

\begin{table}[H]
\centering
\caption{Custom model built for UCF sports action dataset}
\begin{tabular}{|l|l|l|l|l|}
\hline
Training set & Testing set & Iterations & \# Accuracy \\
\hline
Imagenet & UCF Sports Action & 10,000,000 & 71.96\\
UCF Sports Action & UCF Sports Action & 10,000 & 54.43\\
UCF Sports Action & UCF Sports Action & 85,000 & 43.68\\
\hline
\end{tabular}
\label{table:training_difference}
\end{table}

A possible reason the model trained on Imagenet actually outperformed the model trained on 
the UCF sports action dataset could be the result of the reference model being trained 
for a large amount of time with powerful GPU. 
This demonstrates the importance of training the lower level weights to 
better extract information from the input. Over-fitting may have been a 
reason why increasing the iterations by 75k decreased performance. Even though the imagenet model was trained on 10,000,000 iterations, imagenet is enormous, so it would be unlikely that the same image was seen more than once or twice. If over-fitting was a culprint, then for some portion of the final 75,000 training iterations, the CNN was solely learning a random noise distribution unique to the training set.

\subsection{Testing Feature Vector Concatenation}

Initially we assumed that a concatenation of the neuron activations of the $6^{th}$ and $7^{th}$ fully connected layers would be the best. 
We soon realized that the $7^{th}$ layer was very sparse and often
doesn't contain useful information for classification. 
We put together a script to do different concatenations of the feature levels. 
We used the imagenet trained model caffe provided and tested combinations of layers.

\begin{table}[H]
\centering
\caption{F16, F19, F22 feature vector concatenations}
\begin{tabular}{|r|r|r|r|r|}
\hline
Feature Layers & Layer Size Length & J48 decision tree & 3-NN & Poly Kernel SVM\\
\hline\hline
F16 & 4096 & \textbf{60.59} & 65.76 & 69.74\\
F19 & 4096 & 46.12 & 70.58 & 71.76\\
F22 & 1000 & 42.47 & 67.05 & 69.62\\
\hline
F16 $\oplus$ F19 & 8192 & 48.01 & \textbf{71.34} & \textbf{71.96}\\
F16 $\oplus$ F22 & 5096 & 50.94 & 69.67 & 70.50\\
F19 $\oplus$ F22 & 5096 & 44.44 & 70.50 & 71.50\\
\hline
F16 $\oplus$ F19 $\oplus$ F22 & 9192 & 42.47 & 67.05 & 69.62\\
\hline
\end{tabular}
\label{table:vector_cat}
\end{table}

From Table \ref{table:vector_cat} we learn the two following things:
\begin{enumerate}
\item Using a support vector machine is needed for accurate classification. 
We experimented with J48 pruned decision trees to get a very fast solution but it was often about 20\% less accurate as a support vector machine. 
KNN3 was better than decision trees, but once again didn't measure up in accuracy.
\item Intermediate layers (F16 and F19) were more accurate than the final layer (F22). 
This result is not what we expected because the final layer should encapsulate the final classification.
Perhaps it is caused by the F22 Imagenet layer being trained for 1,000 class labels and the UCF dataset containing only 9 classes.
In light of this, it appears that the F22 layer is not going to be helpful.
\end{enumerate}

\subsection{Fine Tuning}

A large portion of the experimentation  on the UCF sports 
action dataset entailed modifying the architecture of the network. 
We left the first 5 convolutional layers (layers 1 -15 in Table \ref{table:architecure}) undisturbed in all of these experiments. 
In layers 1 - 15 there are an enormous number of parameters that can be explored in the future; kernel size, number of kernels, number of layers, pool size, pool stride, etc. 
Based on previous SVM results with F16 concatenated with F19, this was one of the two featurization levels tested. 
Also because we were fine tuning specifically for this data set, we tested the F22 layer for accuracy.

The purpose of fine tuning is to use an already well established model 
(for example the imagenet trained model), and replace the top layers to match the new problem. 
The lower layers are unchanged. 
The theory behind this is then we will remove the dataset specific classifiers from the model, but still retain the well honed low level featurization that is universal to all 
images. 
We swept in size over three levels to experiment for the best size: F16, F19, and F22.

In the F22 layer we tried 9 nodes (since there are 9 classes it is attempting to classify), and 1000 nodes (this is the default F22 layer size of imagenet, but these 1000 nodes need to be trained to discriminate sports action characteristics). 
The results with a variable size F22 layer are shown below in Table \ref{table:finetuning}. 
All of the layer sizes not listed (c1 through ds15, fc16, relu17, d18, fc19, relu20, d21) were unchanged from Table \ref{table:architecure}. 
Unless stated otherwise, the learning rate alpha was set at .0001 (learning sometimes diverged when set higher), and 20,000 iterations were run.

\begin{table}[H]
\caption{Fine tuning of F22 layer}
\centering
\begin{tabular}{|l|l|l|}
\hline
Fine Tuning Level & F16 $\oplus$ F19 feature accuracy & F22 feature 
accuracy\\
\hline
F22 = 9 & 65.46 & 66.01\\
F22 = 500 & 69.36 & 62.56\\
\hline
F22 = 1000 & 69.36 & 62.56\\
\hline
F22 = 2000 & 60.34 & 58.79\\
\hline
\hline
F22 = $1000^\dagger$ & 72.20 & 58.52\\
\hline
\end{tabular}
\label{table:finetuning}
\end{table}
\textit{$\dagger$ = 1,000,000 iterations.} \\

In the first two experiments we ran (F22 = 9, F22 = 1000), F22 was the most 
telling feature vector for the class label. 
We split the difference between the two for the first follow up
experiment (approximately at F22 = 500), and doubled the previous best size 
for a second follow up experiment (F22 = 2000), and was surprised  that in both of these tests the F16$\oplus$F19 feature layer was so much more accurate than the F22 layer.
We took the most accurate configuration of the four trials (F22=1000 nodes), and ran 
them for a long weekend for 1 million iterations. 
The F16$\oplus$F19 concatenated layer had the best performance of any of the other
experiments and significantly better than the F22 layer.

In trying to increase accuracy further, we fine tuned the F16 and F19 
layers (individually), instead of starting from the default weights of the imagenet model. 
Our method for choosing parameters is a by examining different layer sizes and moving towards the best result. 
We can use this best result as our new midpoint, and split into evenly spaced layer levels again.

When we did our F19 experiment, we started with sizes 2048, 4096, and 8192. 
Since 2048 did poorly, we put our new midpoint between the middle of 4096 and 8192 at a 
layer size of 6144. 
We created evenly spaced tests at 5120 and 7168. 
This method can be used to find a reasonably good guess for the best parameters quickly. Results of fine tuning on the F19 layer can be found in table~\ref{table:fine19}. 
The mean accuracy, as well as the length of the feature vector are given for each 
fine tuning level.

\begin{table}[h]
\caption{Fine tuning of F19 feature vector}
\label{table:fine19}
\begin{tabular}{l|l|l|l|l|l|l|}
\hline
\multicolumn{1}{|l|}{Fine-tune level} & \multicolumn{2}{l|}{F16 $\oplus$ 
F19 Feature Vector} & \multicolumn{2}{l|}{F19 Feature Vector} & 
\multicolumn{2}{l|}{F22 Feature Vector} \\ \hline
                                        & Mean\%             & Vector 
                                        Length            & Mean\%          
                                        & Length         & Mean\%          
                                        &Length         \\ \hline
\multicolumn{1}{|l|}{F19 = 2048}         & 64.46                     & 
6144              
& 65.34                  & 2048           & 63.83                  & 1000           
\\
\multicolumn{1}{|l|}{F19 = 4096}         & 70.71                     & 
8192              
& 70.79                  & 4096           & 70.58                  & 1000           
\\ \hline
\multicolumn{1}{|l|}{F19 = 8192}         & 70.00                     & 
12288             & 70.16                  & 8192           & 67.98                  
& 1000           \\
\multicolumn{1}{|l|}{F19 = 6144}         & \textbf{72.30}            & 
10240             & \textbf{72.22}         & 6144           & 
\textbf{72.30}         & 1000           \\
\multicolumn{1}{|l|}{F19 = 5120}         & 70.21                     & 
9216              
& 69.95                  & 5120           & 68.19                  & 1000           
\\ \hline
\multicolumn{1}{|l|}{F19 = 7168}         & 69.29                     & 
11264             & 69.16                  & 71.68          & 70.42                  
& 1000           \\ \hline
\end{tabular}
\end{table}

The best results were obtained with F19 tuning were a layer  
larger than the imagenet model, 6144 nodes. 
Tuning the F19 layer in this way was able to produce a result superior to the imagenet model by .34\%.

The F16 layer was fine-tuned individually  and the results are shown in table~\ref{table:fine16}. 
We initially used 3 layer sizes chosen with the same method above. 
Due to memory constraints of the GPU, we was unable to expand on the F16 
layer upwards past F16 = 8192, and the CPU implementation was far too slow to attempt.

\begin{table}[h]
\caption{Fine tuning of F16 feature vector}
\label{table:fine16}
\begin{tabular}{l|l|l|l|l|l|l|}
\hline
\multicolumn{1}{|l|}{Fine-tuning level} & \multicolumn{2}{l|}{F16 $\oplus$ 
F19 Feature Vector} & \multicolumn{2}{l|}{F16 Feature Vector} & 
\multicolumn{2}{l|}{F22 Feature Vector} \\ \hline
                                        & Mean\%             & Length            
                                        & Mean\%          & Length         
                                        & Mean\%          & Length         
                                        \\ \hline
\multicolumn{1}{|l|}{F16 = 2048}        & 67.01                     & 6144              
& 66.51                  & 2048           & 65.38                  & 1000           
\\
\multicolumn{1}{|l|}{F16 = 4096}        & 69.37                     & 8192              
& \textbf{68.91}         & 4096           & 68.44                  & 1000           
\\ \hline
\multicolumn{1}{|l|}{F16 = 8192}        & \textbf{70.25}            & 
12288             & 68.36                  & 8192           & 
\textbf{73.23}         & 1000           \\ \hline
\end{tabular}
\end{table}

The best results on the UCF sports action database was obtained with F16 twice the size of the imagenet model, 8192 nodes.
Because of time and memory constraints we was unable to increase the size, but it may prove to be more expressive.
Future experiments could be run with a more powerful graphics card.

Two final experiments were run on the UCF sports action dataset in an 
attempt to get the best performance. 
First we trained the imagenet model with fine tuning. 
In the fine tuning all of the layers past the convolutional layers were set to their 
default values,  then we combined the best parameters from the individual fine tuning 
experiments above. 
The results, which were disappointingly low, are listed in 
table~\ref{table:mini-fine}. 
Using the optimized F6 and F7 layer at the same time were not possible with with current GPU memory. 
This is another future experiment that could be run with a larger GPU.

\begin{table}[h]
\centering
\caption{Imagenet finetuning (F16=4096, F19=4096, F22=1000)}
\label{table:mini-fine}
\begin{tabular}{|l|l|l|l|}
\hline
F16 $\oplus$ F19 & F16 & F19 & F22 \\
\hline
61.69 & 59.55 & 63.33 & 64.12 \\
\hline
\end{tabular}
\end{table}

\begin{table}[h]
\centering
\caption{Optimized Architecture finetuning (F16=4096, F19=6144, F22=1000)}
\begin{tabular}{|l|l|l|l|}
\hline
F16 $\oplus$ F19 & F16 & F19 & F22 \\
\hline
63.70 & 63.37 & 61.69 & 62.78 \\
\hline
\end{tabular}
\end{table}

Overfitting was a concern in both of these experiments. 
Part of the CAFFE software allows the printing out of loss (from cross entropy) at intervals of your choosing and the loss usually does not reach zero.
However, we found it reached zero quickly in both of these architectures, which  is indicative of over-fitting. 
Further experimentation should explore why this happened, and more importantly how to fix it. 
We hypothesize that varying the percentages of dropout levels in the $6^{th}$ and $7^{th}$ fully connected levels would enhance performance.

\section{Fine Grained Classification}

In addition to the UCF sports action dataset, six additional datasets were 
used as described above in section~\ref{subsection:datasets}. 
Of the six, three of them can be considered fine grained sets (Olivetti faces, Caltech-UCSB birds, and Stanford dogs), and the other three contain
many more generic classes (Caltech101, Caltech256, VOC2012). 
We mean by "fine grained" that the the classes are subsets of classes of what we would usually 
consider a class in itself. 
That is, instead of a “dog” class, classes in Stanford Dogs set are types of dogs,  such as “Siberian Husky”, or “Yellow Labrador retriever”. 
In Table \ref{table:finegrained}, the first two columns are information about the dataset and the next three are using the pretrained imagenet model on F16, F19, and F22 featurization levels. The next column is the fine tuned model with the same layer sizes as imagenet for all layers except for F22. 
In the final column the model is fine tuned with only layers up to F16. 
The F16 layer output size is set at a constant (5000), because I
hypothesized that a larger output size would be necessary if it were the 
only non-convolutional layer. 
The results are displayed in the table below. 
Due to the size of these datasets, SVMlight was used as opposed to weka for SVM classification. Of the datasets appearing in table~\ref{table:finegrained}, the number classes (top 
to bottom) are: 101, 200, 256, 40, 200, 20; 
the number of F8 output nodes are: 101, 200, 256, 40, 200, 200; 
and the number of F6 output nodes are: 5000, 5000, 5000, 5000, 5000, 5000. 
These were left out of the table due to the relative unimportance of the 
content, and the size and readability constraints of the table.
\begin{table}[h]
\centering
\caption{Fine Grained Classification. NET = Imagenet model, FT = Fine tuned model}
\label{table:finegrained}
\begin{tabular}{|l|l|l|l|l|l|l|}
\hline
Dataset        & Size (GB) & NET (F16)      & NET (F19) & NET (F22) & FT (F16)       & FT (F16 out) \\ \hline
Caltech-101    & .157      & \textbf{.7240} & .5388     & .5371     & .6805          & .6520            \\ \hline
CUB-200 Birds  & .694      & \textbf{.3133} & .2403     & .1562     & .2417          & .1653            \\ \hline
Caltech-256    & 1.2       & \textbf{.5587} & .3842     & .3772     & \textbf{.5587} & .2659            \\ \hline
Olivetti Faces & .053      & .9917          & .4917     & .4833     & \textbf{1.000} & .8000            \\ \hline
Stanford Dogs  & .820      & \textbf{.4696} & .3677     & .2342     & .2277          & .0599            \\ \hline
VOC 2012       & 1.4       & \textbf{.5872} & .5204     & .5459     & .4628          & .2516            \\ \hline
\end{tabular}
\end{table}

Three comparisons were run on the datasets: feature level accuracy comparisons,
imagenet vs. finetuning comparisons, and full model vs. truncated model comparisons.

In the feature level accuracy comparison, we first wanted to compare the
expressiveness between the three fully connected layers. 
The first three columns of results confirm what was seen in the UCF sports action datasets; that F16 is the most expressive layer of the fully connected layers. 
Although above we had found that the most accurate feature vector was the
F16 vector concatenated with the F19 vector, here instead we used the F16 vector by itself to aid
in the comparison to the level that does not have an F19 layer.

The second comparison was between the highly trained but not specialized imagenet
model and the minimally trained yet highly specialized models for the specific datasets.
We expected the specialized models to perform better on the three specialized datasets
(birds, faces, and dogs) because of their specificity. 
Out of the six datasets, 4 of them (Caltech101, Caltech200, Stanford dogs, and VOC2012) had better performance on the imagenet dataset, 1 of them (Caltech256) did equally well on both the imagenet and finetuning set, and 1 of them (Olivetti faces) did better on finetuning. 
The Olivetti dataset got 100\% accuracy when fine tuned, which is somewhat suspicious, especially since it is also by far the smallest dataset.

The third comparison was to determine whether fully trained 23 layer models $16^{th}$ fully
connected layer would be more accurate than truncated 16 layer models. 
Intuitively it would make sense that it does, but the non linearities in the $16^{th}$ and $19^{th}$ layers may add to the expressiveness of the model. 
Because these non linearities exist, a multilayer perceptron system may be able to express more than a single layered perceptron. 
For this reason we hypothesized that datasets with less classes (VOC has 20 classes, Olivetti with 40 classes, Caltech101 with 101 classes) to do better
with a single layer output succeeding the 5 convolutional layers. 
All 6 out of 6 datasets did better on the full size model as opposed to the truncated model even though the F16 layer was the features extracted in both models. 
This suggests that the layers of nonlinearity helps the expressiveness even if the only way that the F16 layer benefits from the nonlinearity is backpropagation.

\section{Conclusion and Discussion}
Though this is a promising start, due to time and resource constraints it is by no means
a complete study of evaluating convolutional neural networks for use. 
The current graphics processing units be used (GTX 480 on the cluster, and GT 730M on the laptop) are inferior to state of the art graphics processing units such as the NVIDIA Tesla
40k, which materially limited this investigation.
Not only does training take longer, less meaningful experiments can be done in
the allotted time and the performance was worse.
In addition, there was not enough GPU memory to execute some of the models described in this study. 

In the Caltech256 dataset, there are over 20,000 images in  
the training set, so not all of the images are even seen in training. 
This was done to make sure that the tests could be completed in a reasonable time. 
In a real world situation the system would see every image many times. 
The only image set that did better on fine tuning had 280 total training images, and at 20,000
iterations, so each training image was seen to be trained on over 71 times. 
The image set that had the largest decrease in performance between the imagenet model and the fine tuned model was the Stanford Dogs dataset. 
There was 13,680 images in the training corpus, so the average image was seen less than 2 times. 
With more computational power we would surely be able to increase the performance.

The outcome of the experiments that were run were promising, and demonstrate the
power of classification in convolutional neural networks. 
We recommend that convolutional neural networks be pursued further in computer vision tasks, based on their accuracy and ways that their training time can be minimized.

This work represents an initial investigation and future work includes leveraging time series of features and voting for the majority label of videos.



\printbibliography
\end{document}